\documentclass[conference]{IEEEtran}
\usepackage{cite}
\usepackage{subcaption}

\usepackage{amsmath,amssymb,amsfonts}
\usepackage{algorithmic}
\usepackage{graphicx}
\usepackage{enumitem}
\usepackage{textcomp}
\usepackage{graphicx,caption}

\usepackage{multicol}
\usepackage{xcolor}
\def\BibTeX{{\rm B\kern-.05em{\sc i\kern-.025em b}\kern-.08em
    T\kern-.1667em\lower.7ex\hbox{E}\kern-.125emX}}
\begin{document}

\title{Measures of Complexity for Large Scale Image Datasets}

\author{\IEEEauthorblockN{Ameet Annasaheb Rahane \textsuperscript{\textsection}} 
\IEEEauthorblockA{\textit{University of California, Berkeley} \\
Berkeley, CA \\
ameetrahane@berkeley.edu}
\and
\IEEEauthorblockN{Anbumani Subramanian}
\IEEEauthorblockA{\textit{Intel}\\
Bangalore, India \\
anbumani.subramanian@intel.com}
}

\maketitle
\begingroup\renewcommand\thefootnote{\textsection}
\footnotetext{This work was done as an intern at Intel.}
\endgroup

\begin{abstract}
Large scale image datasets are a growing trend in the field of machine learning. However, it is hard to quantitatively understand or specify how various datasets compare to each other - i.e., if one dataset is more complex or harder to ``learn'' with respect to a deep-learning based network. In this work, we build a series of relatively computationally simple methods to measure the complexity of a dataset. Furthermore, we present an approach to demonstrate visualizations of high dimensional data, in order to assist with visual comparison of datasets. We present our analysis using four datasets from the autonomous driving research community - Cityscapes, IDD, BDD and Vistas. Using entropy based metrics, we present a rank-order complexity of these datasets, which we compare with an established rank-order with respect to deep learning.
\end{abstract}

\begin{IEEEkeywords}
Machine learning, deep learning, manifold learning, intrinsic dimensionality, learning theory
\end{IEEEkeywords}

\section{Introduction}
With an ever-increasing use and need for large image datasets for machine learning, it is important to see how datasets differ in terms of complexity and distribution. When comparing large scale image datasets, the common practise is to evaluate metrics like  total number of images, number of images in each class, or class distributions in the dataset. However, all these are metrics defined by humans and do not really  provide any insights about the underlying distribution of data. 

Our objective in this work is to enable the comparison of the \emph{unconditioned} distribution of data between any two datasets. In other words -- our interest in the analysis of a comparative complexity of data between different image datasets and to analyze whether this aspect of complexity impacts the ability of a neural network to learn from that dataset. Towards this,  we utilize concepts from statistics, topology, and information theory -- to capture the essence of  data dimensionality without any knowledge of underlying data labels or classes in the dataset. We focus on three main factors and relate these metrics to a network learning scheme. Following are the main contributions of our work:
\begin{enumerate}[label=(\roman*)]
    \item A study on the distribution of entropy for a given dataset
    \item Estimation of the intrinsic dimensionality of a given dataset and thus learn about the dimensions of underlying data manifold
    \item Use of Uniform Manifold Approximation and Projection (UMAP) \cite{UMAP} to transform a given dataset, and thus visually compare the transformations 
\end{enumerate}

In this work, our focus is on the datasets widely used in autonomous driving community, where semantic segmentation of images is a problem of high interest.  Towards this problem, a number of new image datasets have been released in the recent years. These datasets are targeted for research in object detection and  semantic segmentation of images from a vehicle on roads. A good performance on semantic segmentation on image datasets informs the feasibility of building perception algorithms, a critical blocks for self-driving, autonomous vehicles in the future. Therefore, analyzing the kind of complexity in datasets offers a glimpse of how ``hard'' it is to learn from a dataset compared to others and hence provides insights into  building  generalizable agents for the problem of perception in self-driving. Our objective is to build, test and demonstrate a comparison of complexity and distribution of data in large volume, high dimensional image datasets.

\section{Datasets}
In this work, we use four well known datasets in the field of autonomous driving: Cityscapes (CS) \cite{city}, India Driving Dataset (IDD) \cite{IDD}, Berkeley Deep Drive (BDD) \cite{BDD}, and Mapillary Vistas (Vistas) \cite{vistas}. We chose these datasets specifically to study and compare if the nature of datasets have any influence on data complexity and distribution.  The datasets like CS, BDD, or Vistas contain images from structured driving conditions while  IDD was designed with images from unstructured driving conditions and hence our interest here.

\section{Methods}

\subsection{Shannon Entropy}
In our work, we measure entropy in three different ways. The standard Shannon entropy of a grayscale image is defined as:  
\label{shannon_desc}
\begin{equation}
\label{entropy_calc}
H = -\sum_{i=0}^{n-1}p_i \log p_i
\end{equation}
where $n$ is the number of gray levels and $p_i$ is the probability of a pixel having a given gray level $i$. 

\subsection{GLCM}  
\label{glcm_desc}
The gray level co-occurence matrix (GLCM) of an image is another commonly used entropy measure. GLCM is a histogram of co-occuring grayscale values at a given offset over an image. GLCM helps to characterize the texture in an image, beyond the image properties like energy and homogeneity. Here, the GLCM entropy is calculated as:
\begin{equation}
    H_g = -\sum_{i=0}^{n-1}\sum_{j=1}^{n-1}p(i,j)\log p(i,j)
\end{equation}
where $n$ is the number of gray levels and $p(i,j)$ is the probability of two pixels having intensities $i$ and $j$, separated by the specified offset. 

\subsection{Delentropy}
\label{del_desc}
Delentropy is a measure based on a computable probability density function called \textit{deldensity} \cite{del}. This density distribution is capable of capturing the underlying structure using spatial image  and pixel co-occurrence. This is possible because the  scalar image pixel values are non-locally related to the entire gradient vector field, which means that the usage of gradient vectors in the calculation of this entropy allows for global image features to be considered in the calculation of delentropy. This is in contrast to the pixel level Shannon entropy, which entirely relies on individual pixel values. Delentropy  is defined similar to Equation \ref{entropy_calc} but uses  with the new (deldensity) distribution which can capture more non-local information. \cite{del}

\subsection{Intrinsic Dimensionality Estimation}

\label{sec:idest}
Intrinsic Dimensionality (ID) represents the minimal representation of the underlying manifold possible for a dataset without losing any information. In other words, it is the smallest dimension required for a deep learning model to exactly represent a dataset. If $m$ is the intrinsic dimensionality of a dataset, it implies that a $m$ dimensional manifold is required to represent the data in the dataset and that any representation less than $m$ cannot cover the data without any loss of information. 

\subsubsection{Estimator}
The maximum likelihood estimator of intrinsic dimensionality is defined in \cite{id_est}. 

Consider a dataset with $n$ images and let $X_i$ represent the individual sample (image) in the dataset. Let $X_1,\cdots, X_n \in \mathbb{R}^p$ be i.i.d. observations with an embedding of a lower dimensional sample. Let each sample $X_i$ be mapped to a point $x_i$ in the manifold. Specifically, $X_i = g(Y_i)$, where $Y_i$ are sampled from an unknown smooth density $f$ on $\mathbb{R}^m$, with some $m\leq p$, and $g(.)$ is a continuous and sufficiently smooth mapping function. $m$ is the intrinsic dimensionality of the dataset. 

In other words, we pick a point $x$, assuming $f(x)$ is approximately constant in a sphere of radius $R$ around $x$. Then, we can treat the observations as a homogeneous Poisson process in $S_x(R)$. Using the derivation from \cite{id_est} and following the notations, the maximum likelihood estimation for $m$, given $k$ nearest neighbors, is
\begin{equation}
    \label{eqn:mle}
    \hat{m}_k(x) = \Bigg[\dfrac{1}{k-1}\sum_{j=1}^{k-1}\log \dfrac{T_k(x)}{T_j(x)}\Bigg]^{-1}
\end{equation}
where $T(x)$ is the average distance from each point to its $k$-th nearest neighbor. This is an estimator of the intrinsic dimensionality which is shown to balance bias and variance better than all other existing solutions \cite{id_est}.

\subsection{UMAP: Uniform Manifold Approximation and Projection}
We use UMAP \cite{UMAP}, a topology preserving dimensionality reduction technique to create visual representation of the embedding of a dataset, and then qualitatively see how this embedding relates to other metrics (discussed here) and also relate to available performance comparison of deep learning methods.

\subsubsection{Assumptions}
The use of uniform manifold approximation and projection requires a few assumptions. UMAP  assumes that data must be uniformly distributed on a Riemannian manifold, the manifold is locally connected and the Riemannian metric is locally constant or can be approximated as such.

\subsubsection{Embedding}
UMAP provides a manifold learning technique for dimension reduction, with foundations in Riemannian geometry and algebraic topology. The first phase consists of constructing a fuzzy topological representation by using simplices. Then, we optimize our embedding (by using stochastic gradient descent) to have as close a fuzzy topological representation as possible (measured by cross entropy). This enables UMAP to construct a topological representation of the higher dimensional data in a lower dimensional space. \cite{UMAP}

\subsubsection{Method}

In our work, we define a vector space of image as follows. We define a mapping from each image to a vector in a high dimensional space and define some distance function between any two points in that space. The most simple version of this is the space of images with $n$ pixels each, pointing to a vector in $\mathbb{R}^n$ (flattening the image into a vector). However, we can create a more intuitive representation of the image space as a vector space, using the gray level co-occurence matrix (GLCM). Using this feature space, we can find the dimensionality reduction dependent on the topology of texture based features. Using this space, we can assume that the conditions listed above are met.

Using  additional parameters like number of nearest neighbors in UMAP, we show how local and global structures change/shift differently in different datasets - i.e., when the nearest number of neighbors to a point are low, it implies that more local structure is used to create the embedding, and when the number of nearest neighbors to a point are high, higher order structure in the dataset is used to create the embedding. \cite{UMAP}

\subsection{Computation} 
We used the implementation of ID from \cite{intdim_mle_github} and performed all our experiments on an Intel Xeon\texttrademark \hspace{0.05in}  processor. The entire computation for intrinsic dimensionality was completed well within 60 seconds for Cityscapes, IDD and BDD datasets. This computation increased to over an hour for Vistas, which contains lot more images than other datasets.

\section{Results}

\subsection{Entropy}
\begin{figure*}[h]

\minipage{0.25\textwidth}
\includegraphics[width = \linewidth]{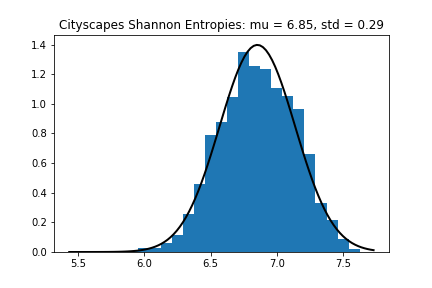}
\endminipage \hfill
\minipage{0.25\textwidth}
\includegraphics[width = \linewidth]{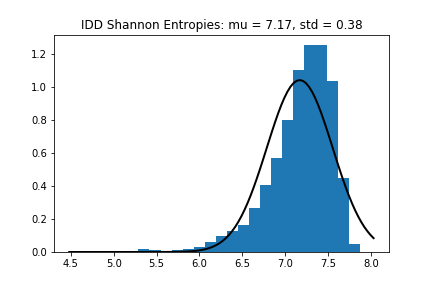}
\endminipage \hfill
\minipage{ 0.25 \textwidth}
\includegraphics[width = \linewidth]{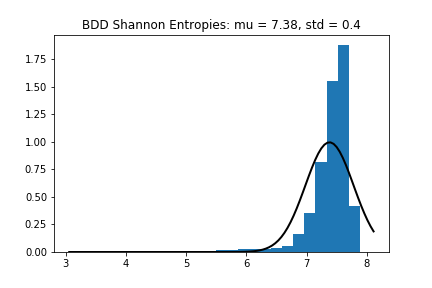}
\endminipage \hfill
\minipage{0.25 \textwidth}
\includegraphics[width = \linewidth]{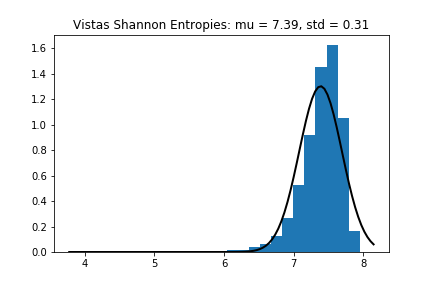}
\endminipage \hfill

\caption{Shannon entropy distributions fit to normal distributions of all four datasets. Notably, as we move left to right from Cityscapes to Vistas, the distribution appears to become much tighter in nature. This can also be seen as the standard deviation becomes smaller from Cityscapes to BDD. This implies that BDD and Vistas have fewer lower entropy images than CS and IDD. As such, when a network learns these datasets, the first few layers of the network will have a harder time learning CS and Vistas.}
\label{shannon}
\end{figure*}

\begin{figure*}[h]

\minipage{0.25\textwidth}
\includegraphics[width = \linewidth]{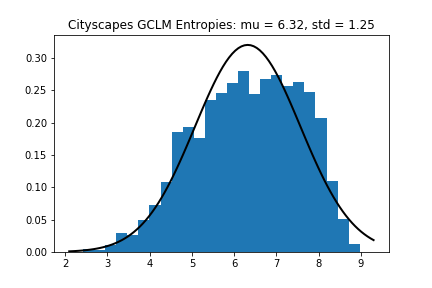}
\endminipage \hfill
\minipage{0.25\textwidth}
\includegraphics[width = \linewidth]{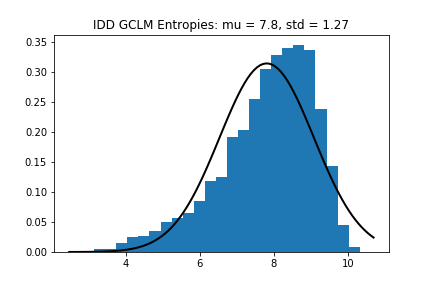}
\endminipage \hfill
\minipage{0.25 \textwidth}
\includegraphics[width = \linewidth]{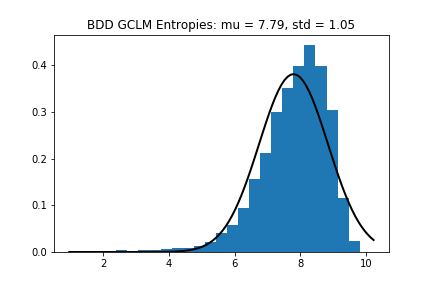}
\endminipage \hfill
\minipage{0.25\textwidth}
\includegraphics[width = \linewidth]{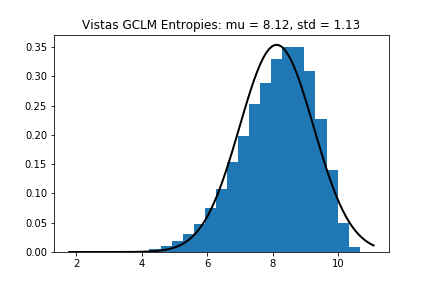}
\endminipage \hfill

\caption{GLCM Entropy distributions fit to normal distributions of all four datasets - Cityscapes, IDD, BDD and Vistas. Contrary to \figurename \ref{shannon}, the texture level entropy of each dataset does not exhibit a shift in notable standard deviation. }
\label{glcm}
\end{figure*}

\begin{figure*}[h]
\minipage{0.25\textwidth}

\includegraphics[width = \linewidth]{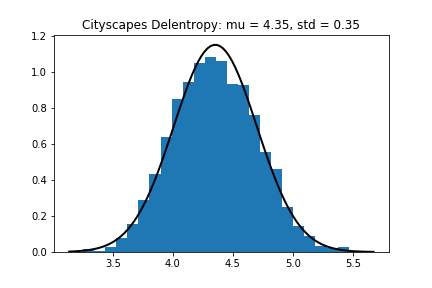}
\endminipage \hfill
\minipage{0.25\textwidth}
\includegraphics[width = \linewidth]{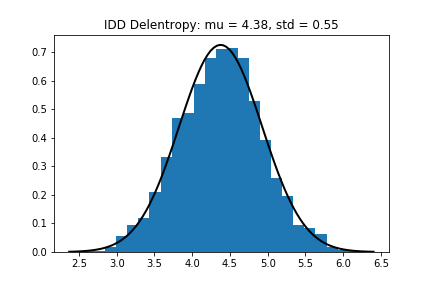}
\endminipage \hfill
\minipage{ 0.25 \textwidth}
\includegraphics[width = \linewidth]{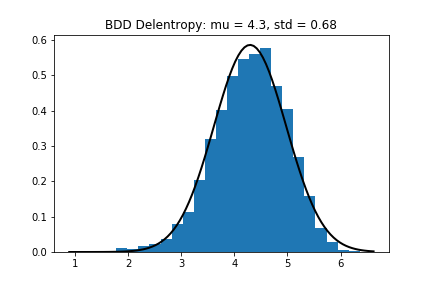}
\endminipage \hfill
\minipage{ 0.25 \textwidth}
\includegraphics[width = \linewidth]{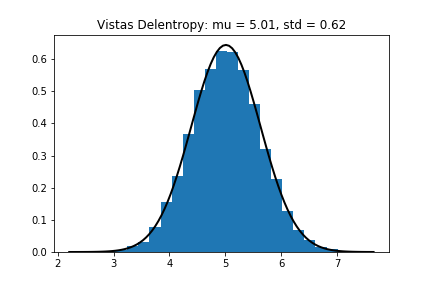}
\endminipage \hfill
\caption{Delentropy distributions fit to normal distributions of all four datasets. Delentropy seems to  be normally distributed. This could be a feature of the deldensity calculation, but this points out that global features do not exhibit a severe shift in entropy between the given datasets.}
\label{del}
\end{figure*}

For Cityscapes, IDD, BDD, and Vistas, we used standard implementations of the methods described in sections \ref{shannon_desc}, \ref{glcm_desc}, \ref{del_desc} to calculate Shannon Entropy, GLCM Entropy, and delentropy for every image in each dataset. In Table \ref{tab:entropy_rank}, we present the average entropy of each dataset (for each type of entropy). figure \figurename \ref{shannon}, \ref{glcm}, and \ref{del} show the entropy distribution in each dataset (fitted to a normal distribution). 

\subsection{Intrinsic Dimensionality Estimation}
Using the implementation from \cite{intdim_mle_github}, we calculate the maximum likelihood estimation of intrinsic dimensionality for each dataset and these results are shown in Table \ref{tab:id}. From this table, we can infer the geometric complexity of our datasets, and also get some interesting insights into the representations of the given datasets.\\

\vspace{2pt}
\begin{table}[h]
    \centering
    \caption{Intrinsic dimensionalities of the selected datasets.}
    \begin{tabular}{c|c}
        Dataset &  Intrinsic Dimensionality (ID)\\
        \hline
         Cityscapes & $11.4$\\
         IDD & $11.9$\\
         BDD & $13.7$\\
         Vistas & $30.1^*$\\
    \end{tabular}
     \\
   \vspace{3pt}{ $*$:  Vistas had a high estimation variance possibly due to the large dataset size, and computational restrictions.
   } 
    \label{tab:id}
\end{table}

\subsection{UMAP: Uniform Manifold Approximation and Projection}
We use UMAP to help visualize the entire datasets in our work. We find an approximation to the topology of GLCM features of each of our datasets, and then use it to embed the four datasets in $\mathbb{R}^2$.  

In \figurename \ref{umap_fig}, we show a plot of each two dimensional embedding. Alongside each axis is the distribution of each dimension of the embedding. Overall, it can be seen as a plot of the distribution of a two-dimensional embedding of the dataset, with the marginals on each axis. We have shown plots of four variations with the number of neighbors $k=2, 20, 100, 500$, to show the change in datasets, considering local to global focused embeddings.

\section{Observations}

\subsection{Entropy}
Of the four datasets used here, Vistas is considered  most difficult to learn for semantic segmentation algorithms. From the entropy plots, we observe possible supporting evidence in a  quantitative manner - Vistas has the highest mean entropy across all entropies, and more so for GLCM based entropy implying that Vistas has relatively more complex texture features than the other three datasets. 

Based on the  entropy distributions, we formulate a  rank order of complexity (lowest to highest) for the four datasets. This rank order is supported by the plots in \figurename \ref{shannon}, \ref{glcm}, and \ref{del} adn the tables in Table \ref{tab:mu_std_entropy} and \ref{tab:entropy_rank}.

\begin{table}[h]
    \centering
    \caption{Mean (and standard deviation) of entropies of the datasets.}
    \label{tab:mu_std_entropy}
    \begin{tabular}{c|c|c|c}
         Dataset & Shannon pixel entropy & GLCM (texture) entropy &  Delentropy\\
         \hline 
        Cityscapes &  $6.85 (0.29) $ & $6.32 (1.25)$ &$4.35 (0.35)$  \\
         IDD& $7.17 (0.38)$ & $7.8 (1.27)$   &  $4.38 (0.55)$ \\
         BDD& $7.38 (0.4) $ & $7.79 (1.05) $&    $4.3 (0.68)$\\
         Vistas& $7.39 (0.31)$ &$8.12 (1.13) $ &  $5.01, (0.62)$ \\
    \end{tabular}
\end{table}

\vspace{2pt}
\begin{table}[h]
    \centering
    \caption{Rank order of datasets with respect to average entropies.}
    \label{tab:entropy_rank}
    \begin{tabular}{c|c|c}
         Shannon pixel entropy & GLCM (texture) Shannon entropy &  Delentropy\\
         \hline 
        Cityscapes & Cityscapes &  BDD \\
         IDD &BDD  &  Cityscapes\\
         BDD &IDD &  IDD \\
         Vistas &Vistas &  Vistas \\
    \end{tabular}
\end{table}

BDD has a higher pixel entropy, although its texture and more global structure are lower than those of IDD. For a deep learning network based semantic segmentation method, a higher pixel entropy implies that the first few layers of the network which interpret pixel distributions, will have difficulty in learning and could manifest as issues like convergence of bias weights, time taken to converge. But for the same network, in case of IDD, the latter layers would have a difficulty in learning. Approximately, high Shannon pixel entropy would imply that a given network would have a harder time learning the first few layers, as these layers are used to interpret pixel level behavior. Alternatively, high delentropy and GLCM texture entropy would mean that the network would have a harder time learning in the middle or later layers, as these layers are used to learn higher order features. 

From our analysis, using an entropy-centric view, we observe that the Cityscapes dataset tends to be the lowest in complexity, aside from delentropy. Delentropy represents a higher order structure of the images. We see that Cityscapes, BDD, and IDD are of  similar complexity with respect to delentropy. This observation can be correlated by the fact that these three datasets were created out of a similar data capture setup - dashcam-style images with view of roads -- with differences in geographies and times of capture. Whereas, the Vistas dataset is created with images from both dashcam-style images and mobile-phone images. This capture setup leads to a totally diverse collection of images, with any scene that looks like a road scene. Therefore, the images in Vistas dataset contain lot more texture patterns - i.e., higher order features, This leads to an increased difference between the entropy values, except for Shannon entropy.   

\subsection{Intrinsic Dimensionality}
The significance of Cityscapes and IDD with similar complexity (in Table \ref{tab:entropy_rank}) can be understood as follows: these datasets have images with similar perspective view of the road scene - possibly due to the similarity in data capture setup and camera placements on the vehicle. Moreover,  Cityscapes and IDD lack variations in weather conditions, unlike BDD or Vistas. Considering these factors, it is intuitive to understand the similar pixel based entropy of these two datasets. The difference in entropy is completely dependent on the nature of data in a dataset - in this case, as different landscapes, road patterns, and driving patterns. The object (class) labels of IDD and BDD are compatible with Cityscapes, and so have no impact in the analysis.

\subsection{UMAP}
\label{UMAP_description}
\begin{figure}[h!]
\label{fig:glcm}
\centering

\begin{subfigure}{\textwidth}
\minipage{0.23\textwidth}
\includegraphics[width = \linewidth]{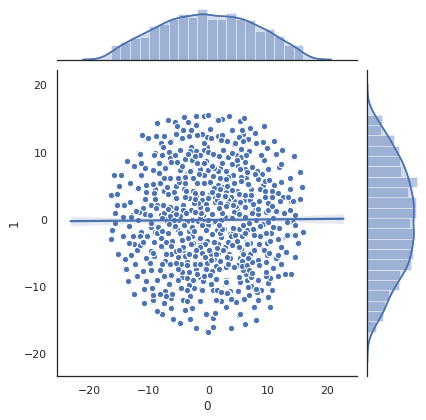}
\endminipage \hfill
\minipage{0.23\textwidth}
\includegraphics[width = \linewidth]{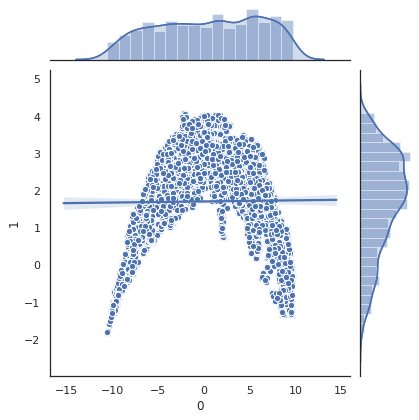}
\endminipage \hfill
\minipage{ 0.23 \textwidth}
\includegraphics[width = \linewidth]{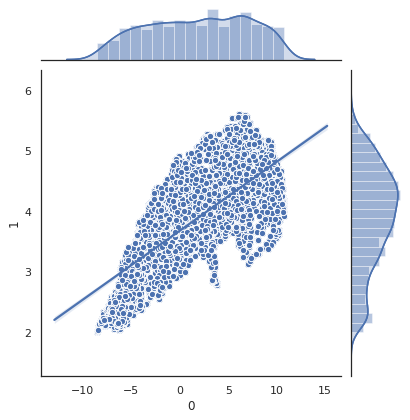}
\endminipage \hfill
\minipage{ 0.23 \textwidth}
\includegraphics[width = \linewidth]{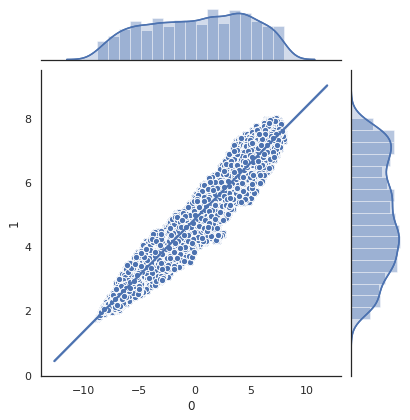}
\endminipage \hfill
\caption{Cityscapes}
\end{subfigure}

\begin{subfigure}{\textwidth}

\minipage{0.23\textwidth}
\includegraphics[width = \linewidth]{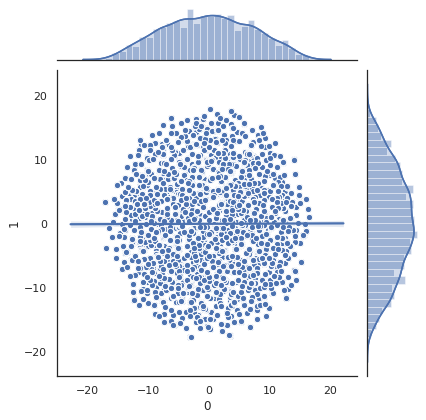}
\endminipage \hfill
\minipage{0.23\textwidth}
\includegraphics[width = \linewidth]{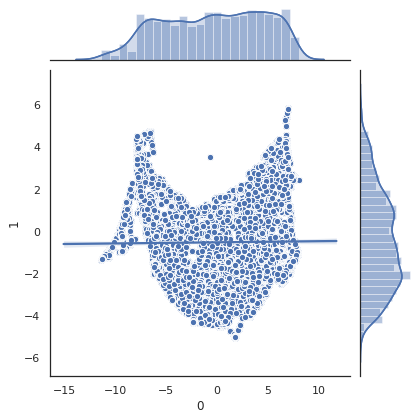}
\endminipage \hfill
\minipage{ 0.23 \textwidth}
\includegraphics[width = \linewidth]{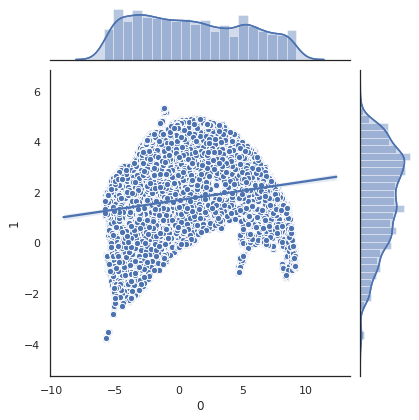}
\endminipage \hfill
\minipage{ 0.23 \textwidth}
\includegraphics[width = \linewidth]{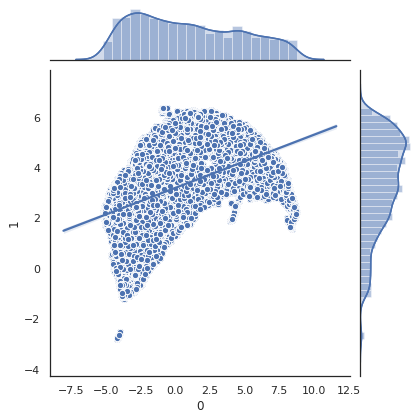}
\endminipage \hfill
\caption{IDD}
\end{subfigure}

\begin{subfigure}{\textwidth}

\minipage{0.23\textwidth}
\includegraphics[width = \linewidth]{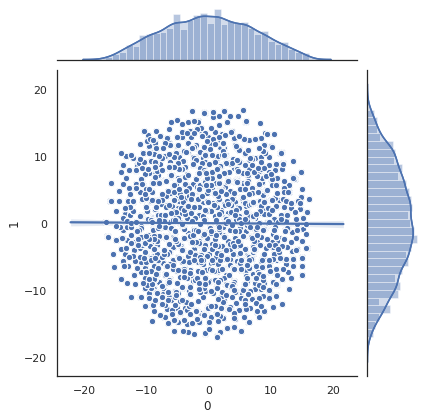}
\endminipage \hfill
\minipage{0.23\textwidth}
\includegraphics[width = \linewidth]{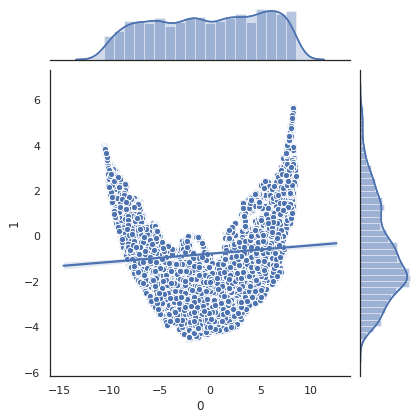}
\endminipage \hfill
\minipage{ 0.23 \textwidth}
\includegraphics[width = \linewidth]{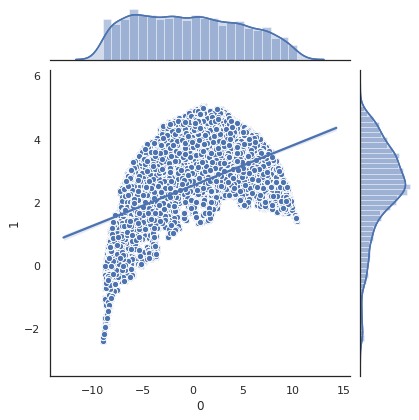}
\endminipage \hfill
\minipage{ 0.23 \textwidth}
\includegraphics[width = \linewidth]{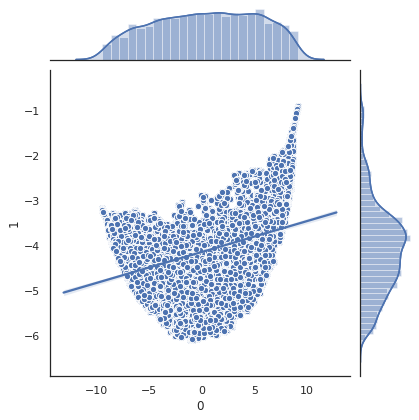}
\endminipage \hfill
\caption{BDD}
\end{subfigure}

\begin{subfigure}{\textwidth}
\minipage{0.23\textwidth}
\includegraphics[width = \linewidth]{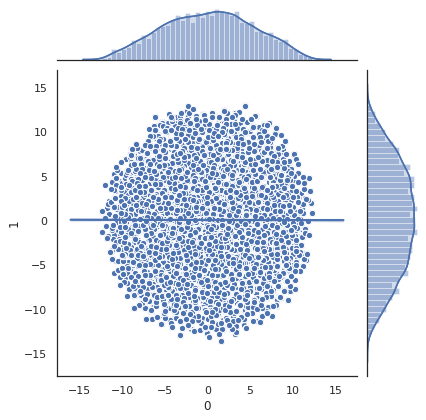}
\endminipage \hfill
\minipage{0.23\textwidth}
\includegraphics[width = \linewidth]{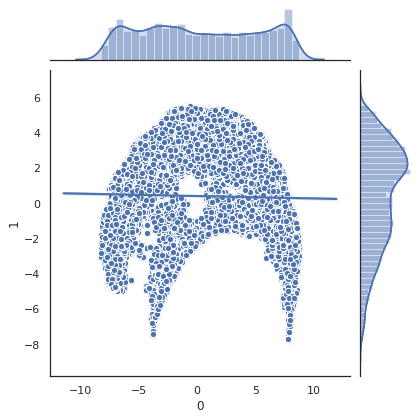}
\endminipage \hfill
\minipage{ 0.23 \textwidth}
\includegraphics[width = \linewidth]{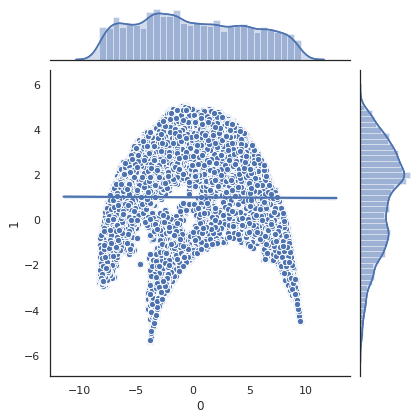}
\endminipage \hfill
\minipage{ 0.23 \textwidth}
\includegraphics[width = \linewidth]{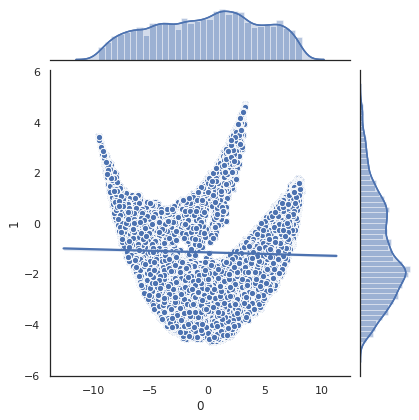}
\endminipage \hfill
\caption{Vistas}
\minipage{0.23\textwidth}

\footnotesize{Neighbors: $2$, Min. Distance: $.1$}
\endminipage \hfill
\minipage{0.23\textwidth}
\footnotesize{Neighbors: $20$, Min. Distance: $.1$}
\endminipage \hfill
\minipage{ 0.23 \textwidth}
\footnotesize{Neighbors: $100$, Min. Distance: $.1$}
\endminipage \hfill
\minipage{ 0.23 \textwidth}
\footnotesize{Neighbors: $500$, Min. Distance: $.1$}
\endminipage \hfill
\end{subfigure}
\\

\minipage{\textwidth}
\hspace{1cm}
    \caption{Two dimensional projection of each dataset using UMAP (Sec. \ref{UMAP_description}). The plots show embedding (with each data point being a coordinate), with the distribution of each dimension on each axis. Many patterns can be be observed, when the embeddeings change from local to global. }
    \label{umap_fig}
\endminipage

\end{figure}

UMAP embedding helps to visualize the projection of each of our datasets. From our visualization plots in Fig. \ref{umap_fig}, we observe that ``difficult'' datasets appear to be more densely distributed. The three datasets - Cityscapes, IDD, and BDD share a lot of similarities in distribution, with minor variations which is easier to observe at more nubmer of nearet neighbors, $k$ (= $100, 500$) (see \figurename \ref{umap_fig}). 

Our observations are consistent with the premise that distribution of data in these datasets have an approximately complexity order, as given in Table \ref{tab:id}: Cityscapes, IDD, BDD, Vistas.

\subsection{Accuracy of deep learning networks}
From the performance metric in \cite{IDD}, we can observe that deep learning semantic segmentation algorithms already have an order of complexity as follows: Cityscapes, IDD, BDD, Vistas. This order matches the rank-order described in our analysis here, and help conclude that pixelwise Shannon entropy, delentropy, intrinsic dimensionality, and deep learning results are related. Given the similarity of entropy distributions, a correlation between delentropy and GLCM feature based entropy apppears to exist.  Using the UMAP specific density plots for network comparison, it is difficult to infer the complexity of a dataset given the higher density of points in the embedding, although it does seem as though a relation exists. These require a more deeper study in future.

\section{Conclusion}
In this work, we demonstrated that well known information theoretic measures like  pixelwise Shannon entropy, GLCM feature Shannon entropy, delentropy, and intrinsic dimensionality all correlate with known results of performance metrics in semantic segmentation networks trained on the same datasets. Therefore, we present these metrics  to further study relationships between information theoretic claims we can make about large scale image datasets and deep learning performance. This could lead lead to providing guarantees about the performance of deep learning on a given dataset or a data generation technique. Additionally, we present a dimensionality reduction embedding in the context of large image datasets to show the density of data clusters as a measure of complexity.

\section{Acknowledgements} Ameet Rahane was supported by the Intel AI Student Ambassadorship program towards this publication.

\clearpage
\newpage

\clearpage

\centering

\end{document}